# Mapping Images with the Coherence Length Diagrams

A. Sparavigna[a] and R. Marazzato [b,c]
[a] Dipartimento di Fisica, Politecnico di Torino, Torino, Italy
amelia.sparavigna@polito.it
[b] Dipartimento di Automatica ed Informatica, Politecnico di Torino, Torino, Italy
[c] Freie Universität Bozen, Bolzano, Italy
roberto.marazzato@polito.it

**Abstract:** Statistical pattern recognition methods based on the Coherence Length Diagram (CLD) have been proposed for medical image analyses, such as quantitative characterization of human skin textures, and for polarized light microscopy of liquid crystal textures. Further investigations are here made on   image maps originated from such diagram and some examples related to irregularity and anisotropy of microstructures shown. The possibility of generating a defect map of the image is also proposed.

## 1. Introduction

Coherence length is a well-known concept in optics and condensed matter physics. In optics, it is the propagation distance from a coherent source till points where the electromagnetic wave maintains a specified degree of coherence [1]. In non-linear optics, coherence length is used to represent the distance of phase-matching between waves. In condensed matter physics, it is the distance over which a specific order is maintained [2]. We translated this concept in an algorithm for image processing analysis, which compares the local and the average behavior of the image map. Let us remember that the image map is a bi-dimensional function of grey/color tones.

The algorithm, based on coherence length analysis, was applied for characterization of irregular textures, where Fourier or wavelets approaches are less useful [3,4]. The algorithm evaluates the smallest area which can exhibits the same grey-tone statistical distribution as that of the whole image. The diagram of coherence lengths gives the boundary of this area [5-7]. Comparing the neighborhood of each pixel with this area, the algorithm is able to determine defective pixels in the image frame.

In our last paper [5], we applied the method to skin characterization. Previously, it was applied to the characterization of liquid crystal textures [6,7]. The main feature of such tool resides in generating the characteristic diagram, the Coherence Length Diagram (CLD), and some maps related to the input image. Here we propose a discussion and several examples: the goal is to explain the nature and some properties of CLD and of four fundamental maps, which can be generated from it: the Support Map (SMap), the Defect Map (DMap), and the Directional Defect Map (DDMap) and the Mixed Map (MMap). CLD and DMap are here presented as the discrete version of the corresponding quantities appearing in [5].

## 2.  The Coherence Length Diagram and its support map.

All computational results we obtain arise from a given image. We only consider the grayscale color coding, and convert other color encoding, such as RGB, when needed. A bitmap representation of an image consists of a function, which yields the brightness of each point within a specific width and height range:

$$b: D \to B \quad (1)$$

with

$$D = I_h \times I_w, \text{ where } I_n = \{1,2,\ldots,n\} \subset \mathbb{N} \quad (2)$$

and

$$B = \{0,1,\ldots,255\} \subset \mathbb{N} \quad (3)$$

$\mathbb{N}$ is the set of natural numbers. The average brightness of an image corresponds to

$$M_o = \frac{1}{hw} \sum_{i=1}^{h} \sum_{j=1}^{w} b(i,j) \quad (4)$$

Consider now an arbitrary point $(i,j) \in D$, as the local origin for an orthogonal reference system and a set of $n_d = 32$ directions

$$\theta_k = k \frac{2\pi}{32} \; ; \; k = 1,\ldots,32 \quad (5)$$

By summing along a specific direction $\theta_k$ up to a certain distance $\lambda$ from $(i,j)$, the local first order moment

$$M_{0,\lambda}^k (i,j) = \frac{1}{\lambda} \sum_{r=0}^{\lambda} b(i + \Delta i_k, j + \Delta j_k) \quad (6)$$

is obtained, which is a function of $\lambda$. In (6), we define $\Delta i_k = [r\cos\theta_k], \Delta j_k = [r\sin\theta_k]$.
For a chosen threshold value $\tau$, we define the local coherence length, if it exists:

$$l_{0,k}^\tau (i,j) = \min\left(\left\{\lambda : \frac{|M_{o,\lambda}^k(i,j) - M_o|}{M_o} \leq \tau\right\}\right) \quad (7)$$

Please note that for some points this quantity is not defined; let us call

$$D_k^\tau = \{(i,j) \in D : \exists \, l_{0,k}^\tau(i,j)\} \subseteq D \quad (8)$$

the subset of $D$ in which the local coherence length over the direction $k$ can be computed for a threshold $\tau$. The local coherence diagram is formed by all (existing) values of coherence length associated to a certain point, considered as a function of the direction index $k$. In order to obtain a diagram which takes into account the coherence length of all points of the given image, the average of $l_{0,k}^\tau(i,j)$ over $D$ is to be computed:

$$l_{0,k}^{\tau} = \frac{\sum_{D_k^{\tau}} l_{0,k}^{\tau}(i,j)}{\text{card}(D_k^{\tau})} \qquad (9)$$

An easy and meaningful way to graphically represent this function is the polar diagram shown in Fig.1.

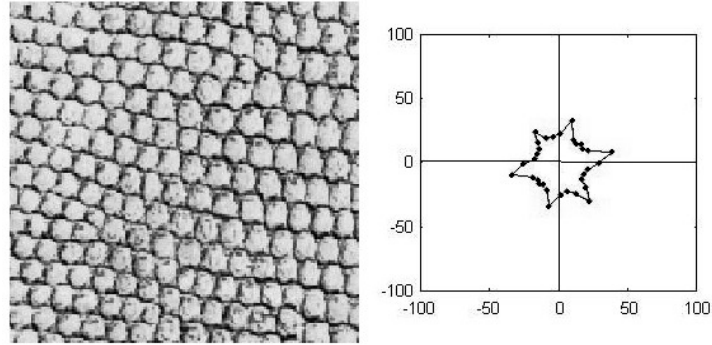

Fig.1 The original image map on the left. On the right, the coherence length diagram evaluated for a threshold $\tau$ of 30%. The CLD is able to indicate the texture anisotropy.

In Fig.1 we see the original image map and its Coherence Length Diagram, or CLD. We used as a first example of CLD evaluation, an image of a snake skin, displayed in the Brodatz Album [8], to visualize the ability of our diagram in revealing preferred directions in the image frame. The texture of image map in Fig.1 is quite anisotropic. The coherence length diagram on the right is clearly depicting the presence of anisotropy in the image and the local behavior of the texture. However, the use of coherence length diagram is more interesting when the image does not contain regular textures, but when images have scarcely regular textures and are quite disordered with several scattered objects within, as those observed in the microscopy investigation of certain nematic liquid crystal cells. [6,7]. Another interesting field of application for CLD is the analysis of average fibers distribution and of grain structures of polycrystalline materials. The maps are shown in Fig.2, with, on the right, the corresponding coherence length diagrams.

Let us note than for a certain threshold value and for each direction $\theta_k$ a support set $D_k^{\tau}$ can be defined. Alternately, this subset can be described through its indicator function [9-11]:

$$\phi_k^{\tau} : D \to \{0,1\}$$
$$\phi_k^{\tau}(i,j) = \begin{cases} 1 & (i,j) \in D_k^{\tau} \\ 0 & (i,j) \notin D_k^{\tau} \end{cases} \qquad (10)$$

This is a well defined and meaningful tool, but representing all image maps arising from 32 different directions leads to almost unreadable results.

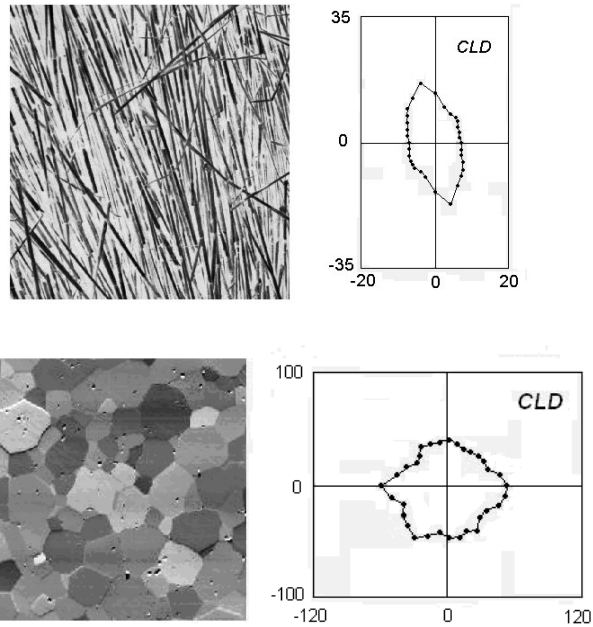

Fig.2 The original image maps are on the left. The upper image is a fibers distribution from the Brodatz Album. The lower image is an optical micrograph of the grain structure of polycrystalline magnesiowüstite. On the right, the coherence length diagram evaluated for a threshold $\tau$ of 30%.

Instead, the average indicator function is used:

$$\phi^\tau : D \to \{0,1\}$$
$$\phi^\tau = \frac{1}{32} \sum_{k=1}^{32} \phi_k^\tau (i,j) \qquad (11)$$

The last expression, called the image Support Map (SMap), is less detailed yet better understandable than the set of single direction support maps. Fig.3 shows a sample of such map, obtained by laying on the given grayscale image a layer, in which the value of the average function is represented by the brightness of the added blue component.

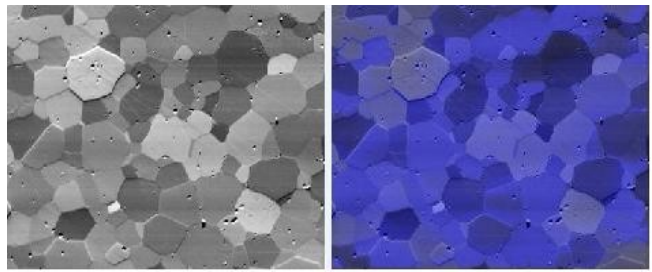

Fig.3 The original image and its support map. Note that there are regions where there were difficulties in evaluating the coherence length, for instance the image boundaries or too bright regions.

## 3. The detection of defects by means of a Defect Map (DMap)

As stated in previous sections, both overall and local coherence diagrams are computed when describing an image. If a comparison between each point's diagram and the CLD is made, possible out-of-average behaviors can be detected for some points. The technique which can be used is quite similar to regular gray level methods [10], but applied to the couple $\{l^{\tau}_{0,k}(i,j), l^{\tau}_{0,k}\}$ instead of $\{b(i,j), M_0\}$. For each point $(i,j)$ and each direction $\theta_k$ such that $(i,j) \in D^{\tau}_k$, the difference between the local and the average coherence length is evaluated, then compared to the threshold $\tau'$, in order to define the directional success function:

$$\psi^{\tau,\tau'}_k : D \rightarrow \{0,1\}$$

$$\psi^{\tau,\tau'}_k(i,j) = \begin{cases} 1 & l^{\tau}_{0,k}(i,j) \in \Delta^{\tau,\tau'}_k \\ 0 & l^{\tau}_{0,k}(i,j) \notin \Delta^{\tau,\tau'}_k \end{cases} \quad (12)$$

where

$$\Delta^{\tau,\tau'}_k = \left[\, l^{\tau}_{0,k}(1-\tau'),\, l^{\tau}_{0,k}(1+\tau')\,\right]. \quad (13)$$

Now the defect map (DMap) can be defined as

$$\psi^{\tau,\tau'} : D \rightarrow \{-1,1\}$$

$$\psi^{\tau,\tau'}(i,j) = 2 \frac{\sum\limits_{k:(i,j)\in D^{\tau}_k} \psi^{\tau,\tau'}_k(i,j)}{\sum\limits_{k=1}^{32} \phi^{\tau}_k(i,j)} - 1 \quad (14)$$

which basically counts the number of "correct" local values. The DMap is displayed (see Fig.4) by adding an RGB layer and encoding positive values of to $\psi^{\tau,\tau'}(i,j)$ to green additional components, and negative values to red ones. CLD and DMaps are also useful in the case of images with regular pattern.

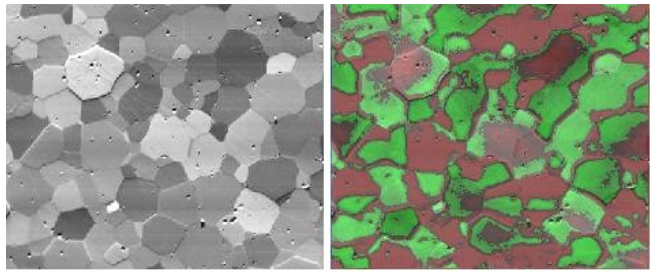

Fig.4 The DMap of grain distribution is displayed on the right. It is obtained by adding an RGB layer to the original map. The threshold value $\tau'$ if 50%.

In the following Fig.5, it is possible to see a chessboard with a missing black site. The CLD diagram displays the regular features of the map. The related Dmaps are also shown for three different values of the threshold τ′: a suitable choice of this value is able to identify the position of defects in the image texture.
Many other examples can show the versatility of Dmaps. Among them we choose a uniform surface with several dots on it, with the proper threshold to enhance the background and dot position (see Fig.6). The background gray tone is practically the reference level: the Dmap pinpoints the defects, which are the dots on the surface.

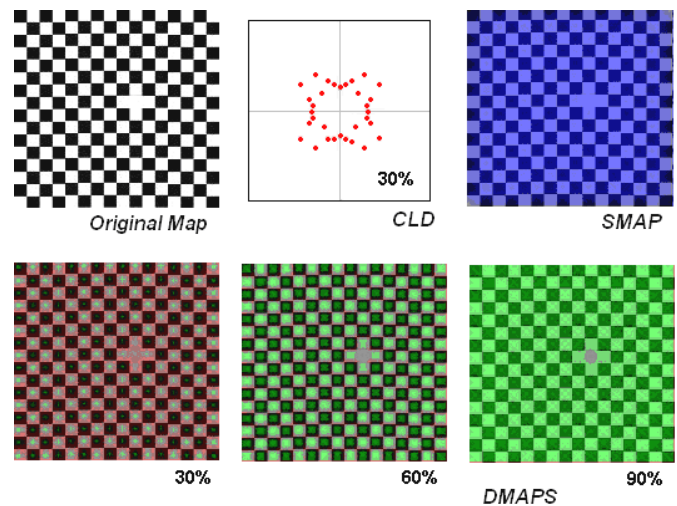

Fig.5 A chessboard has a defect. This defect does not change the CLD diagram (evaluated for a threshold τ of 30%). Comparing the local neighborhood of each pixel with the CLD, the DMaps are obtained for three different value of the threshold τ′. In the DMap on the right, the position of defect is highlighted.

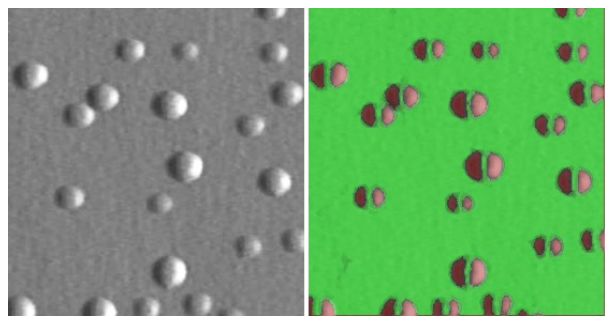

Fig.6. A uniform surface with several spots on it. With the proper threshold we enhance the background, dot positions and their features (bright or dark sides).

## 4. The Directional Defect Map (DDMap).

The Defect Map described in previous section discriminates between points behaving "almost like" and "definitely unlike" the average CLD, but it is not focused on shape differences. A shape comparison can be made by using a square difference analysis involving the local and the average coherence length diagram. The sum of square differences

$$Q^\tau(i,j) = \sum_{k:(i,j)\in D_k^\tau} \left(l_{0,k}^\tau(i,j) - \overline{l_{0,k}^\tau}\right)^2 \qquad (15)$$

should be compared to its overall average in order to define a corresponding defect map for a given acceptance threshold $\tau''$. However, a scaling issue occurs: in facts, if for certain points the local and the average diagram show a similar shape but different sizes, then a "large" square difference is found where a "small" one should be.
Thus, a scale factor

$$\rho^\tau(i,j) = \frac{\sum_{k=1}^{32} \phi_k^\tau(i,j) \sum_{k=1}^{32} \overline{l_{0,k}^\tau}}{32 \sum_{k:(i,j)\in D_k^\tau} l_{0,k}^\tau(i,j)} \qquad (16)$$

is considered. Moreover, the number of actual directions involved in the previous sum affects the value of such quadratic sum, so that the result must be referred to that value. This is represented by the factor

$$\sigma^\tau(i,j) = \frac{32}{\sum_{k=1}^{32} \phi_k^\tau(i,j)} \qquad (17)$$

appearing in next formula. Under these conditions, the normalized sum is obtained:

$$\widetilde{Q}^\tau(i,j) = \sigma^\tau(i,j) \sum_{k:(i,j)\in D_k^\tau} \left(\rho^\tau(i,j) l_{0,k}^\tau(i,j) - \overline{l_{0,k}^\tau}\right)^2 \qquad (18)$$

then the local value $\widetilde{Q}^\tau(i,j)$ is compared to its average value

$$\langle \widetilde{Q}^\tau \rangle = \frac{1}{hw} \sum_{i=1}^{h} \sum_{j=1}^{w} \widetilde{Q}^\tau(i,j) \qquad (19)$$

in order to eventually define the Directional Defect Map (DDMap), given by

$$\delta^{\tau,\tau''} : D \to \{0,1\}$$
$$\delta_k^{\tau,\tau''}(i,j) = \begin{cases} 1 & \widetilde{Q}^\tau(i,j) \in H^{\tau,\tau''} \\ 0 & \widetilde{Q}^\tau(i,j) \notin H^{\tau,\tau''} \end{cases} \qquad (20)$$

where

$$H^{\tau,\tau''} = \left[\ <\tilde{Q}^\tau>(1-\tau''), <\tilde{Q}^\tau>(1+\tau'')\ \right] \qquad (21)$$

Fig.7 shows how a layer is superimposed to the original image, which renders "one" values as yellow added components. Note that the map is able to find the boundaries of the domains.

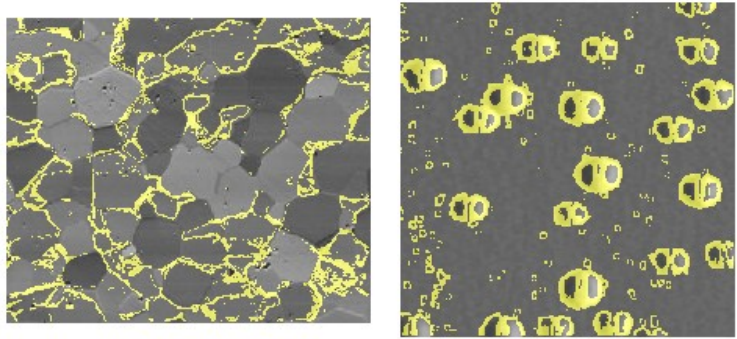

Fig.6 DDMaps of images in Fig.2 and Fig.6. The images are obtained with a superimposition to the original image, which renders the "one" values from Eq.20 as yellow pixels. Note that the maps are able to find the boundaries of domains.

### 5. Mixing the maps in the Mixed Map (MMap)

The previously described algorithms have been tested on some reference geometrical shapes, on some standard images taken from the Brodatz archive [8] and on some camera and microscope pictures of mineral structures. What can be noticed is that DMaps outline success and defect areas, while DDMaps stress boundaries of both sharply and smoothly defined image parts. This different behaviour arises from DDMaps sensing the orientation of local CLDs, which show sudden changes and well-defined directions at boundaries of shapes. A very useful mapping of both features is obtained with a Mixed Map (MMap), where the two layers, DMap and DDMap, are superimposed to the original image (Fig.8). MMaps report boundaries separating the domains with different features. Let us note that the image features are strictly connected with the choice of thresholds. We will discuss the role of such choice in a future publication.

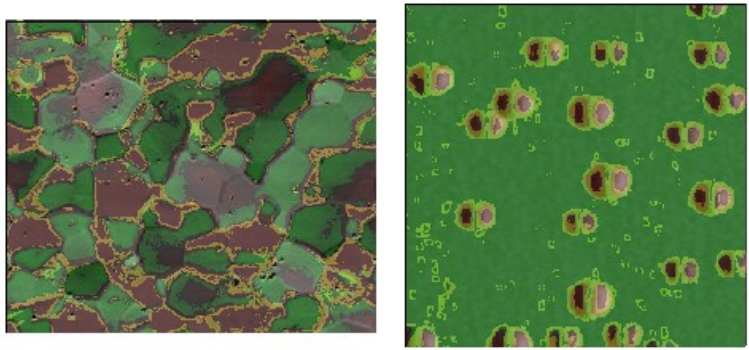

Fig.8 MMaps of images in Fig.2 and Fig.6. The images are obtained with a superimposition of DMap and DDMap to the original images. Note the boundaries which separate the domains with different features.

## 6. Conclusions

The paper describes discrete algorithms based on the Coherence Length Diagrams. With these diagrams it is possible to introduce a defect map (Dmap) which is able to outline defective areas. Another map, the directional defect map (DDMap) stresses the boundaries of both sharply and smoothly defined image parts. This different behavior arises from the fact that the DDMap is sensing the orientation of local CLDs, which shows sudden changes as well as defined directions at boundaries of shapes. In fact, the DDMap is an improvement with respect to algorithms for the simple edge detection.

## Acknowledgements

Many thanks to Florian Heidelbach, Bayerisches Geoinstitut, Universität Bayreuth for providing images of mineral structures to be processed, as the optical micrograph of the grain structure of polycrystalline magnesiowüstite in its undeformed state shown in Fig.2 [12].